# A Ground Segmentation Method Based on Point Cloud Map for Unstructured Roads

Zixuan Li, Haiying Lin, *Zhangyu Wang, Huazhi Li, Miao Yu and Jie Wang

## ABSTRACT

Ground segmentation, as the basic task of unmanned intelligent perception, provides an important support for the target detection task. Unstructured road scenes represented by open-pit mines have irregular boundary lines and uneven road surfaces, which lead to segmentation errors in current ground segmentation methods. To solve this problem, a ground segmentation method based on point cloud map is proposed, which involves three parts: region of interest extraction, point cloud registration and background subtraction. Firstly, establishing boundary semantic associations to obtain regions of interest in unstructured roads. Secondly, establishing the location association between point cloud map and the real-time point cloud of region of interest by semantics information. Thirdly, establishing a background model based on Gaussian distribution according to location association, and segments the ground in real-time point cloud by the background substraction method. Experimental results show that the correct segmentation rate of ground points is 99.95%, and the running time is 26ms. Compared with state of the art ground segmentation algorithm Patchwork++, the average accuracy of ground point segmentation is increased by 7.43%, and the running time is increased by 17ms. Furthermore, the proposed method is practically applied to unstructured road scenarios represented by open pit mines.

*Z. Wang (✉)
Research Institute for Frontier Science, Beihang University, Beijing 100083, China
State Key Lab of Intelligent Transportation System, the Key Laboratory of Autonomous Transportation Technology for Special Vehicles, Ministry of Industry and Information Technology. Beijing 100191
e-mail: zywang@buaa.edu.cn

Z. Li · H. Lin · H. Li ·J. Wang
School of Transportation Science and Engineering, Beihang University, Beijing 100083, China

M. Yu
School of Mechanical Electronic & Information Engineering,China University of Mining and Technology(Beijing), Beijing, 100083, China

# INTRODUCTION

With the continuous development of autonomous driving environment perception technology, point cloud processing technology has attracted much attention. Point cloud processing technology can express objects in three-dimensional space in the form of point clouds. The semantic information obtained from point cloud map plays an important role in the field of autonomous driving environment perception.

In the application of point cloud processing technology, ground segmentation is a basic task. Ground segmentation can help self-driving vehicles detect and identify foreground objects more accurately, thereby improving their driving safety and work efficiency. However, the current ground segmentation algorithms [1, 2, 3] perform poorly in unstructured road environments, such as the limitation of under-segmentation.

Inspired by roadside lidar-based background segmentation methods, which achieved by subtracting background map built from static point clouds from real-time point clouds [4]. Therefore, this paper utilizes the point cloud map constructed by the vehicle lidar to be subtracted from the real-time point cloud to realize ground segmentation. However, there are many challenges in this method. Firstly, the shape of the boundaries are irregular and traffic signs rarely appear in the unstructured road, which leads to difficulty to extract the region of interest (ROI) when the ground is segmented. Secondly, the coordinate system of the map point cloud and real-time point cloud are different and the vehicle shaking produced by poor road conditions, which will seriously affect the performance of ground segmentation. Thirdly, the point cloud dense map can find the corresponding position through pixels and semantics, and the information of real-time point cloud is sparse, leading to difficult for ground segmentation of point cloud.

Considering that the prior background in the point cloud map contains rich semantic information, this paper takes advantage of the semantic information of the point cloud map to solve the above problems. Firstly, the ROI extraction module roughly divides the real-time point cloud into ground point cloud and non-ground point cloud according to the boundary semantic information of the point cloud map, and provides data support for point cloud registration. Secondly, the point cloud registration module optimizes the relative position parameters between the map and the real-time point cloud via the ground and non-ground semantic information of the map within the ROI. Thirdly, the background substraction module establishes the background model and segments the ground point via the relative position parameters. Finally, the evaluation in real applications demonstrates the effectiveness of the ground segmentation method.

The main contributions of this paper are summarized as follows:

(1) **A ground segmentation framework** based on point cloud map that ground data segmented in real-time point cloud by processing semantic information of the map.

(2) **Extracting regions of interest** based on semantic association, which solves the problem of difficult region extraction due to irregular road boundaries.

(3) **A point cloud registration method** based on map semantic information, which realizes the data association between map point cloud and real-time point cloud and reduces the impact of vehicle shaking on ground segmentation performance.

# 1. RELATED WORK

The current mainstream ground segmentation methods include methods based on grid maps, methods based on plane models, and methods based on background image background substraction.

The method based on the grid map mainly divides the grid in different areas and divides the ground according to the threshold in the grid [5, 6, 7]. Literature [5] establishes a grid map based on x - y and segments the ground point cloud according to the height threshold of non-ground points and ground points. Literature [6] realizes segmentation according to the height change threshold. Literature [7] establishes a grid which fits a straight line model through discrete points in the form of polar coordinates, and the ground point cloud is segmented according to the distance threshold. The raster map-based method is prone to mis-segmentation problems in unstructured scenes, and the threshold cannot be adjusted according to the environment, resulting in poor accuracy.

The method based on the planar model mainly constructs a planar model and extracts ground points [8, 9, 10, 11, 12]. Literature [8] employs the random sampling consistency method (RANSAC) to construct a plane model and optimize the plane fitting parameters by filtering out constant values. Literature [9] of the random sampling consensus algorithm is considered to filter out outliers, but the long time of fitting the model affects the accuracy of ground segmentation. Literature [10] proposed the ground plane fitting algorithm (GPF), which screens the fitting points and speeds up the fitting speed. But the ground segmentation accuracy is not ideal. Literature [11] proposed Patchwork, based on the model of concentric circles, and the ground is segmented according to the normal vector of the ground, which still contains insufficient segmentation problems. Literature [12] proposed Patchwork ++ to alleviate the under-segmentation problem through ground likelihood estimation (GLE), but the accuracy performed in unstructured roads is still not good.

The method based on the map background substraction mainly establishes the background model through the static point cloud, and segments the background point cloud in the real-time point cloud via the background substraction method [13,14,15,16,17]. Literature [13] build a background model for the voxel height and points of the roadside point cloud, and segment the ground by height difference. Literature [14] construct the background of the azimuth of the roadside point cloud, and segment the ground by azimuth distance difference. Literature [15] extracts the static ground point cloud from the background substraction in the roadside point cloud according to the high-precision map. In literature [16], the high-difinition map is considered as a prior information, and the ground is extracted the laser point cloud histogram. In Literature [17], the real-time ground data is segmented according to the height difference after registering the local map with the real-time point cloud. Compared to the roadside lidar, the relative position of the vehicle lidar and the ground changes frequently, accompanied by the shaking of the vehicle body, so the ground cannot be directly extracted by the background substraction method. Moreover, compared with the background required for roadside point cloud background substraction, the map required for vehicle point cloud occupies more memory, which is also a challenge in terms of real-time performance.

To sum up, methods based on raster maps and methods based on planar models cannot adapt to complex environments, accompanied by problems such as mis-segmentation and under-segmentation. In order to adapt to the complex environment of unstructured roads, this paper proposes a method based on map background substraction for ground segmentation of point cloud data.

## 2. GROUND SEGMENTATION ALGORITHM

### 2.1 Algorithm Overview

The algorithm framework is shown in Fig. 1. The input are raw data from lidar and point cloud map. The algorithm as a whole can be divided into three parts. (1) ROI extraction. Considering the complexity of the unstructured road environment, the boundary is not clear and there is no sequence and it is impossible to directly utilize the boundary to extract the exercisable area. Therefore, this paper proposes a method based on image processing. The steps include: establishing semantic association to obtain the boundary point cloud in the map, creating a grid map according to the boundary, floodingfill the grid map, extracting ordered boundary points, and extracting real-time ROI from the cloud. (2) Point cloud registration. As a result of the turbulence of the vehicle, point cloud registration is required to ensure the location accuracy from the prior map. Therefore, this paprer proposes a point cloud registration method. Under the condition that the map ground is a plane constraint, the transformation relationship of the local coordinate system is optimized, and then the relative position of the map and the real-time point cloud in the ROI is optimized under the condition of the parameters. (3) Background substraction. The background substraction includes: establishing the background model of the registered map based on the Gaussian model, subtracting in background of the real-time point cloud based on the statistical threshold and updating background model afterwards.

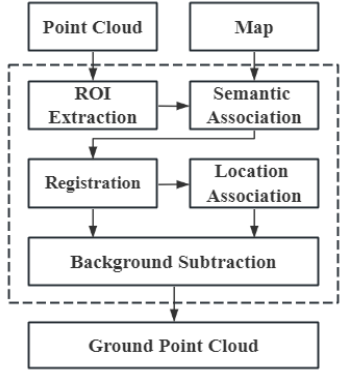

**Fig. 1** Algorithm Overview

### 2.2 Region of Interest Extraction

Since the unstructured road boudaries extracted from map are unclear and disordered, it is difficult to extract the ROI directly. Therefore, it is necessary to sort the boundary points and calculate the ROI by polygon area extraction. Considering that the road within the boundary is a closed area, the ROI inside the boundary of the point cloud can be converted into a connected domain filling of the image contour for processing.

The ROI extraction process is shown in Fig. 2. First, the boundaries in the map are obtained by semantic association, and then the boundaries converted into a rasterized image. Then, flooding fill ROI, and the contours are finded. Finally, contour points converted to polygon points for computing interior area.

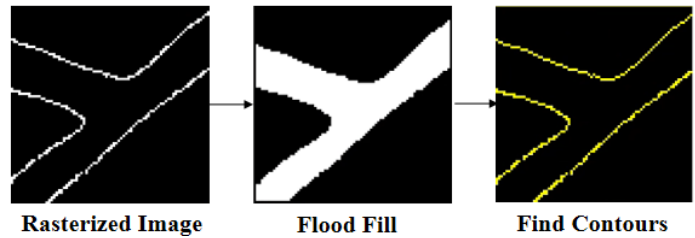

**Fig. 2** Overview of ROI Extraction Method

The boundary semantics of the road boundary points are obtained through the point cloud map, and the boundary semantics of 60 meters around the vehicle body are searched the nearest neighbor search algorithm, and the boundary points in the point cloud map are extracted.

The grid map is based on the vehicle position as the origin, 50 meters ahead, 10 meters behind, and 50 meters left and 50 meters right to create a step length of 1 meter. The boundary points near the vehicle are drawn into the map, and the R value, G value, and B value are respectively taken as Relative to the original position information of the vehicle's lateral offset x, longitudinal offset y, and height offset z. The color-filled grid is considered to create a masked image based on the boundary information.

The flood filling algorithm that filling the grid with color from the center of the vehicle to the outside. (1) Select the pixel where the vehicle is located as the seed point. (2) With the seed point as the center, the pixels whose pixel values of the eight neighbors differ from the pixel value of the seed point by less than the threshold join the area. (3) the currently added pixels as new seed points, and repeat (2) until no new pixels are added to the region.

Extracting the position of the boundary point is to find the grid with the order of polygons obtained by the contour search algorithm, and restore the position information of the boundary point according to the color of the grid and the position information stored in RGB. Finally, the point cloud within the ROI is extracted based on the boundary points with polygon order.

According to the method of literature [18], the shrinkage or expansion of the boundary can be calculated to new ROI.

## 2.3 Point cloud registration

Due to the difference between the vehicle body coordinate system during mapping and the real-time detection vehicle body coordinate system and the unavoidable lidar shaking in the complex road environment, the prior background provided by the point cloud map and the real-time detection background cannot be completely overlapped. In order to overcome these problems, a point cloud registration method is proposed in this paper, which is developed based on the semantic information of the point cloud map.

In the grid map, extract the ground point cloud $S_1 = \{s_i\}_{i=1,2,3,...,N_1}$ in the point cloud map within the 1.5-meter area of the indented boundary and the non-ground point cloud $S_2 = \{s_i\}_{i=1,2,3,...,N_2}$ outside the area of interest, and extract the real-time point cloud within the 1.5-meter area of the indented boundary Ground point clouds $G_1 = \{g_j\}_{j=1,2,3,...,M_1}$ as well as non-ground point clouds $G_2 = \{g_j\}_{j=1,2,3,...,M_2}$ outside the ROI.

According to the Levenberg-Marquardt optimization algorithm of literature [19], the positional relationship $[t_z, \theta_{roll}, \theta_{pitch}]$ between the map point cloud $S_1 = \{s_i\}_{i=1,2,3,...,N_1}$ and the real-time point cloud is calculated by the property that the ground point $G_1 = \{g_j\}_{j=1,2,3,...,M_1}$ cloud conforms to the surface features. Then, as a constraint, traverse the grid of occupied boundary points, and calculate the height mean $[t_z, \theta_{roll}, \theta_{pitch}]$ and variance $\sigma^2$ of the real-time point cloud in each grid's 24 neighborhoods, and $z_0$ take the grid with the largest variance and its 8 neighborhood grids. Take the map point cloud in the grid area as the source point cloud $S_2 = \{s_i\}_{i=1,2,3,...,N_2}$, and the real-time point cloud in the grid area as the target point cloud $G_2 = \{g_j\}_{j=1,2,3,...,M_2}$, assuming that the map point cloud and real-time point cloud in the grid area satisfy the normal distribution:

$$s_i \sim N(\hat{s}_i, C_i^S) \quad (1)$$

$$g_j \sim N(\hat{g}_i, C_j^G) \quad (2)$$

where $C_i^S$ and $C_j^G$ are the covariance matrices of the source point cloud and the target point cloud in their respective neighborhood sets, respectively. The distance between the point cloud of each point in the source point cloud and the target point cloud after passing through the transformation matrix:

$$d_i' = \sum_j (g_j - Rs_i) \quad (3)$$

The above distances satisfy a normal distribution:

$$d_i' \sim (\sum_j (g_j - Rs_i), C_j^G + RC_i^S R^T) \quad (4)$$

According to the method of maximum likelihood to iteratively solve the optimal transformation matrix into

$$R = \arg\min_R \sum_i N_i (\frac{\sum g_j}{N_i} - Rs_i)^T (\frac{\sum C_j^G}{N_i} + RC_i^S R^T)^{-1} (\frac{\sum g_j}{N_i} - Rs_i) \quad (5)$$

$$R = [t_x, t_y, \theta_{yaw}] \quad (6)$$

where is the number of points in the target point cloud in the neighborhood of the source point cloud, $N_i$ and $\frac{\sum C_j^G}{N_i}$ the distribution sum of the points closest to the source point cloud $\frac{\sum g_j}{N_i}$ is a constant.

Finally, the registered map point cloud is obtained according to the registration result:

$$S_{opt} = RS \tag{7}$$

### 2.4 Background Substraction

the real-time point cloud. Therefore, the background model is established according to the Gaussian distribution of the registered map point cloud:

$$P(S_{opt}) = \frac{1}{\sqrt{2\pi}\sigma} e^{\frac{(z-z_0)^2}{2\sigma^2}} \tag{8}$$

After registration, the height values of point clouds in each grid in the map point cloud conform to a normal distribution, where and are the mean and variance of the heights of all point clouds in each grid, $z_0$ respectively $\sigma^2$.

For each point on the real-time point cloud in the ROI, $(x, y)$ the $[y]$ calculation is performed $i=[x]$ in the corresponding background grid, where $(i, j)$, $j=[y]$ and $[x]$ represent $i$ the largest integers not greater than and respectively. $j$ like:

$$|G(x, y) - S_{opt}(i, j)| > k\sigma \tag{9}$$

Wherein is the specified threshold, then this point is the foreground point. if :

$$|G(x, y) - S_{opt}(i, j)| \leq k\sigma \tag{10}$$

Then this point is the background point, and the mean and variance of the background model are updated to obtain the latest background model:

$$S_{opt_t}(i, j) = pG_{t-1}(x, y) + (1-p)G_t(x, y) \tag{11}$$

where $p$ take 0.01.

## 3. EXPERIMENT AND ANALYSIS

### 3.1 Datasets

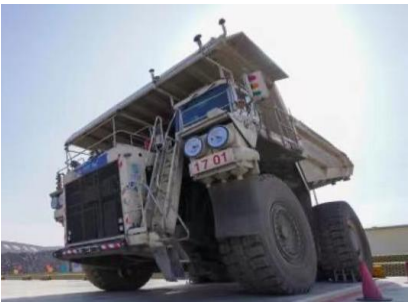
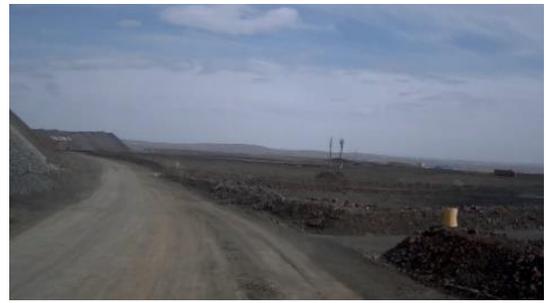

**Fig. 3** Data Collection Vehicle    **Fig. 4** Test site

In order to verify the performance of the algorithm proposed in this paper, this paper collects unstructured road data in an open pit mine in Inner Mongolia. The collection vehicle is shown in Fig. 3. The test site is shown in Fig. 4.

The data set has a total of 12406 frames of data, and the vehicle has traveled a total of 11 kilometers. The mining truck for data collection equipped with the hardware system. The hardware system in this paper mainly includes lidar and integrated navigation. The lidar uses the RS-LIDAR-M1 solid-state lidar of Sagitar Technology, and the scanning frequency is 10Hz. The integrated navigation adopts the CGI-610 integrated navigation system of Huace Integrated Navigation, which supports the output of 10Hz RTK positioning information that accuracy is 1cm.

### 3.2 Segmentation evaluation indicators

The evaluation of segmentation accuracy refers to the methods in statistics [20,21]. Sensitivity represented by $R_{TP}$ and specificity represented by $R_{FP}$. The specific method is

$$R_{TP} = N_{TP} / (N_{TP} + N_{FN}) \qquad (12)$$

$$R_{FP} = N_{FP} / (N_{FP} + N_{TN}) \qquad (13)$$

where $N_{TP}$ is the number of ground points that are correctly marked. $N_{FN}$ is the number of ground points that are wrongly marked as non-ground points. $N_{FP}$ is the number of non-ground points that are wrongly marked as ground points. $N_{FN}$ is the number of non-ground points that are correctly marked. The larger it is $R_{TP}$, the more the number of ground points are correctly segmented, and the better the segmentation effect. On the contrary, $R_{FP}$ the smaller it is, the better the effect. At the same time, the average time spent on processing one frame is calculated as the frame rate indicator.

### 3.3 Operating Experiment

The computer configuration in this experiment is i7-8700K processor, GTX 1080Ti graphics card and 32GB memory. The system environment is Ubuntu 18.04. ROS-medoic, PCL1.8 and OpenCV3.2.0 has been configured. Visual output is implemented by C++ and RVIZ.

The point cloud map in this experiment is a self-developed map whose positioning information with RTK provided by integrated navigation, and point cloud information provided by the Ouster-64 lidar, which is generated by a self-developed mapping algorithm. The point cloud map is shown in Figure 5, the white part is the ground area and the blue part is the non-ground area.

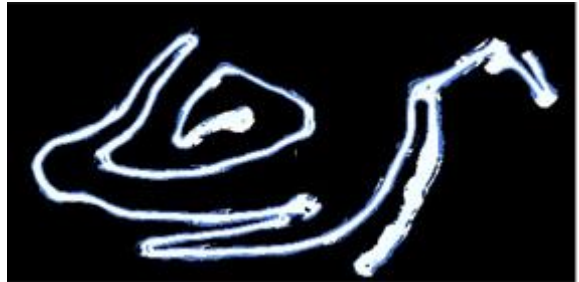

**Fig. 5** Point cloud map

### 3.4 Experimental Results

In order to verify the effectiveness of each module design, the background substraction module is considered as the baseline, and the ROI extraction module and the point cloud registration module are added for experimental verification.

The experimental results are shown in table 1, the data shows that adding the ROI extraction module on the basis of the background substraction module, the sensitivity of the ground segmentation is increased by 0.99 %, and the specificity of the ground segmentation is reduced by 3.13 % . In addition, the performance of ground segmentation is significantly improved after adding the point cloud registration module on the basis of the background substraction module, in which the sensitivity is increased by 8.17%, and the specificity is reduced by 10.14 %. In addition, the ROI is added on the basis of the background substraction module after the extraction module and the point cloud registration module, the performance of ground segmentation is further improved, in which the sensitivity is increased by 15.6%, and the specificity is reduced by 11.38%. The above results prove that the ground segmentation method based on the point cloud map has good results in open pit mines.

Table 1 Ablation experiment results.

| Background Substraction | Region Extraction | Point Cloud Registration | $R_{TP}$/% | $R_{FP}$/% | Time /ms |
|---|---|---|---|---|---|
| √ | | | 84.35 | 16.63 | **16** |
| √ | √ | | 85.34 | 13.50 | 20 |
| √ | | √ | 92.52 | 6.49 | 30 |
| √ | √ | √ | **99.95** | **5.25** | 26 |

The visualization results of the ROI extraction experiment are shown in Fig. 6. The white point cloud is the real-time point cloud, the green point cloud is the boundary semantics in the point cloud map, the red point cloud is the contours of the extracted ROI, and the pink point cloud is the real-time point cloud in the ROI.

The visualization results of the point cloud registration experiment are shown in Fig. 7. The red point cloud is the nearby map point cloud, and the green point cloud is the real-time point cloud.

The visualization results of the ground segmentation experiments are shown in Fig. 8. The white point cloud is the real-time point cloud, the green point cloud is the ground point cloud segmented in the ROI, and the red point cloud is the non-ground point cloud segmented in the ROI.

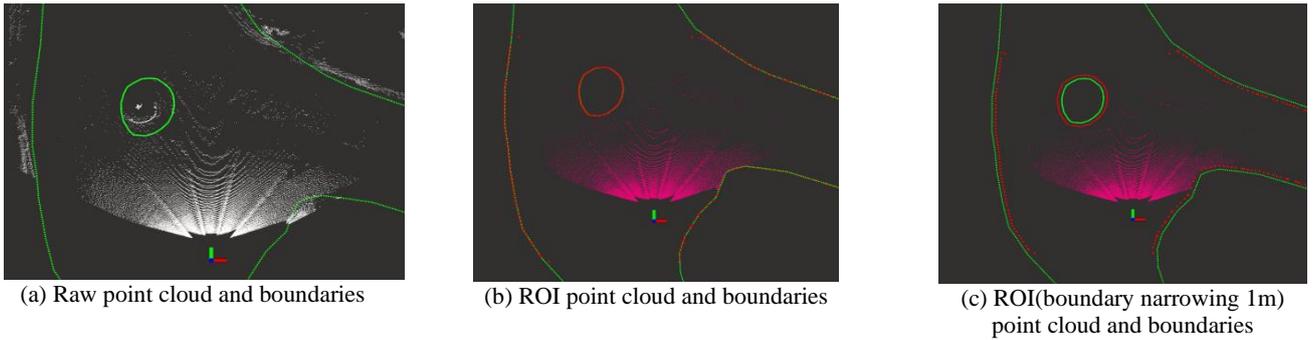

(a) Raw point cloud and boundaries  (b) ROI point cloud and boundaries  (c) ROI(boundary narrowing 1m) point cloud and boundaries

**Fig. 6** The visualization results of the ROI extraction algorithm in the intersection scene

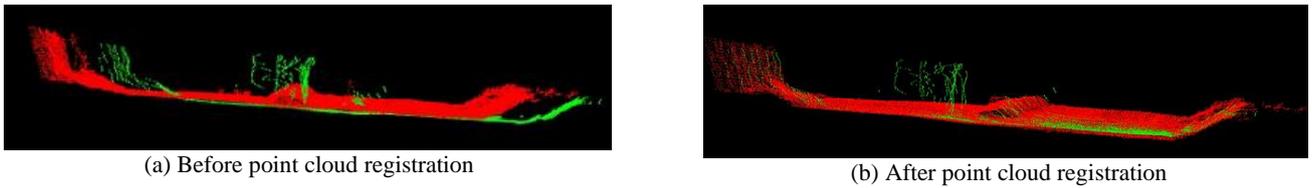

(a) Before point cloud registration  (b) After point cloud registration

**Fig. 7** Visualization results of point cloud registration

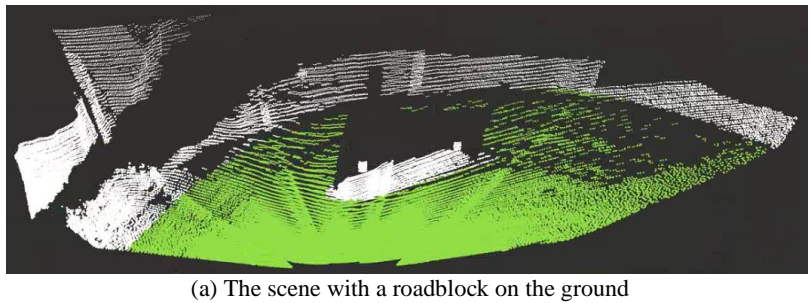

(a) The scene with a roadblock on the ground

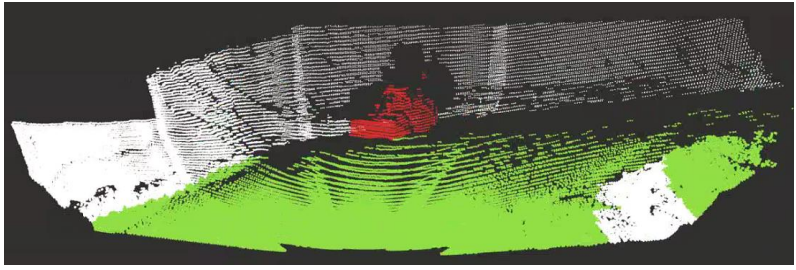

(b) The scene with a power shovel on the ground

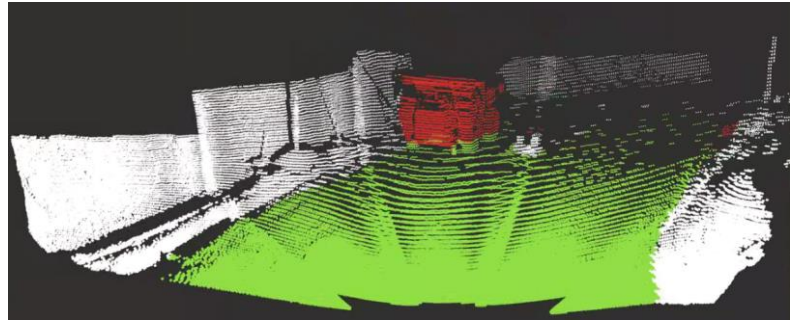

(c) The scene with a mining truck on the ground

**Fig. 8** Visualization results of ground segmentation

In addition, this paper conducts experiments on the mine data set by comparing other ground segmentation algorithms to verify the effectiveness of the proposed algorithm. The experimental results are shown in Table 2. The segmentation accuracy of other methods is lower than the method in this paper, because the method in this paper references the point cloud map and the semantic information of the map makes the results closer to the real value.

**Table 2** Comparative experiment results.

| Algorithm | $R_{TP}$/% | $R_{FP}$/% | Time /ms |
|---|---|---|---|
| Literature [22] | 77.39 | 22.59 | 14 |
| Literature [23] | 90.12 | 11.70 | **5** |
| Literature [11] | 90.88 | 8.93 | 6 |
| Literature [12] | 92.52 | 7.94 | 9 |
| Ours | **99.95** | **5.25** | 26 |

## 4. CONCLUSION

This paper proposes a ground segmentation method based on point cloud map. The sensitivity of ground segmentation is 99.95 %, the specificity is 5.25 % and the average time consumption is 26 ms, which meets the real-time requirements and has been obtained in practical applications confirmed. In addition, the validity of the non-ground point cloud extracted by the method is verified, which not only helps to improve the efficiency of object detection tasks, but also helps to improve the stability and safety of autonomous driving technology. However, the algorithm consumes plenty of time in calculating the nearby map data when extracting the ROI, and the height threshold set in the background substraction is large, resulting in loss of non-ground data. In the future, further analysis will be conducted on these special points to improve the ground segmentation performance.

## REFERENCES

[1] Gomes T, Matias D, Campos A, et al. A survey on ground segmentation methods for automotive LiDAR sensors[J]. Sensors, 2023, 23(2): 601.

[2] Huang SY, Liu LM, Dong J, et al. Review of ground filtering algorithms for vehicle LiDAR scans point cloud data[J]. Opto-Electronic Engineering, 2020, 47(12): 190688.
[3] Liu B, Yu Y, Jiang S. Review of advances in LiDAR detection and 3D imaging[J]. Opto-Electronic Engineering, 2019, 46(7): 190167.
[4] Zhang T, Jin P J. Roadside lidar vehicle detection and tracking using range and intensity background subtraction[J]. Journal of advanced transportation, 2022, 2022.
[5] CHO J, PARK J, BAEK U, et al. Automatic parking system using background subtraction with CCTV environment international conference on control, automation and systems (ICCAS 2016)[C]//2016 16th International Conference on Control, Automation and Systems (ICCAS). IEEE, 2016: 1649-1652.
[6] ALABSI HRH, DEVARAJ JDD, SEBASTIAN P, et al. Vision-based automated parking system[C]//10th International Conference on Information Science, Signal Processing and their Applications (ISSPA 2010). IEEE, 2010: 757-760.
[7] GUO Konghui, LI Hong, SONG Xiaolin. Study on path tracking control strategy of automatic parking system[C]. China Journal of Highway and Transport, 2015, 28(9): 110-118.
[8] Li Mengdi, Jiang Shengping, Wang Hongping. Robust Point Cloud Plane Fitting Method Based on Random Sampling Consensus Algorithm[J]. Science of Surveying and Mapping, 2015, 40(1): 102-106.
[9] Li Xi, Han Xie, Xiong Fengguang. Point cloud plane fitting based on RANSAC and TLS[J]. Computer Engineering and Design, 2017, 38(1): 123-126.
[10] Zermas D, Izzat I, Papanikolopoulos N. Fast segmentation of 3d point clouds: A paradigm on lidar data for autonomous vehicle applications[C]//2017 IEEE International Conference on Robotics and Automation (ICRA). IEEE, 2017: 5067-5073.
[11] Lim H, Oh M, Myung H. Patchwork: Concentric zone-based region-wise ground segmentation with ground likelihood estimation using a 3D LiDAR sensor[J]. IEEE Robotics and Automation Letters, 2021, 6(4): 6458-6465.
[12] Lee S, Lim H, Myung H. Patchwork++: Fast and Robust Ground Segmentation Solving Partial Under-Segmentation Using 3D Point Cloud[C]//2022 IEEE/RSJ International Conference on Intelligent Robots and Systems (IROS). IEEE, 2022: 13276 -13283.
[13] Wang G, Wu J, Xu T, et al. 3D vehicle detection with RSU LiDAR for autonomous mine[J]. IEEE Transactions on Vehicular Technology, 2021, 70(1): 344-355.
[14] Zhang Z, Zheng J, Xu H, et al. Automatic background construction and object detection based on roadside LiDAR[J]. IEEE Transactions on Intelligent Transportation Systems, 2019, 21(10): 4086-4097.
[15] Nagy S, Rövid A. 3D Object Detection in LIDAR Point Cloud Based on Background Subtraction[C]//The First Conference on ZalaZONE Related R&I Activities of Budapest University of Technology and Economics 2022. Budapest University of Technology and Economics, 2022: 20- twenty three.
[16] Ravi Kiran B, Roldao L, Irastorza B, et al. Real-time dynamic object detection for autonomous driving using prior 3d-maps[C]//Proceedings of the European Conference on Computer Vision (ECCV) Workshops. 2018: 0-0.
[17] Muro S, Yoshida I, Hashimoto M, et al. Moving-object detection and tracking by scanning LiDAR mounted on motorcycle based on dynamic background subtraction[J]. Artificial Life and Robotics, 2021, 26(4): 412-422.
[18] Chen X, McMains S. Polygon offsetting by computing winding numbers[C]//International Design Engineering Technical Conferences and Computers and Information in Engineering Conference. 2005, 4739: 565-575.
[19] Ranganathan A. The levenberg-marquardt algorithm[J]. Tutoral on LM algorithm, 2004, 11(1): 101-110.
[20] Zhang Kai, Yu Chunlei, Zhao Yali, et al. Research on Ground Segmentation Algorithm of 3D Laser Point Cloud Based on Adaptive Threshold[J]. Automotive Engineering, 2021, 43(7): 1005-1012.
[21] Rummelhard L, Paigwar A, Nègre A, et al. Ground estimation and point cloud segmentation using spatiotemporal conditional random field[C]//2017 IEEE Intelligent Vehicles Symposium (IV). IEEE, 2017: 1105-1110.
[22] Schnabel R, Wahl R, Klein R. Efficient RANSAC for point‐cloud shape detection[C]//Computer graphics forum. Oxford, UK: Blackwell Publishing Ltd, 2007, 26(2): 214-226.
[23] Arya Senna Abdul Rachman A. 3D-LIDAR multi object tracking for autonomous driving: multi-target detection and tracking under urban road uncertainties[J]. 2017.